**AI agents may be worth the hype but not the resources (yet): An initial exploration of machine translation quality and costs in three language pairs in the legal and news domains**

VICENT BRIVA-IGLESIAS
vicent.brivaiglesias@dcu.ie
Dublin City University

GOKHAN DOGRU
gokhan.dogru@upf.edu
Pompeu Fabra University

Abstract

Large language models (LLMs) and multi-agent orchestration are touted as the next leap in machine translation (MT), but their benefits relative to conventional neural MT (NMT) remain unclear. This paper offers an empirical reality check. We benchmark five paradigms, Google Translate (strong NMT baseline), GPT-4o (general-purpose LLM), o1-preview (reasoning-enhanced LLM), and two GPT-4o-powered agentic workflows (sequential three-stage and iterative refinement), on test data drawn from a legal contract and news prose in three English-source pairs: Spanish, Catalan and Turkish. Automatic evaluation is performed with COMET, BLEU, chrF2 and TER; human evaluation is conducted with expert ratings of adequacy and fluency; efficiency with total input-plus-output token counts mapped to April 2025 pricing.

Automatic scores still favour the mature NMT system, which ranks first in seven of twelve metric-language combinations; o1-preview ties or places second in most remaining cases, while both multi-agent workflows trail. Human evaluation reverses part of this narrative: o1-preview produces the most adequate and fluent output in five of six comparisons, and the iterative agent edges ahead once, indicating that reasoning layers capture semantic nuance undervalued by surface metrics. Yet these qualitative gains carry steep costs. The sequential agent consumes roughly five times, and the iterative agent fifteen times, the tokens used by NMT or single-pass LLMs.

We advocate multidimensional, cost-aware evaluation protocols and highlight research directions that could tip the balance: leaner coordination strategies, selective agent activation, and hybrid pipelines combining single-pass LLMs with targeted agent intervention.

1. Introduction

In recent years, the field of machine translation (MT) has witnessed great advancements. The widespread adoption of neural machine translation (NMT) architectures, particularly those based on the encoder-decoder paradigm, has significantly enhanced translation fluency and adequacy across a variety of language pairs and domains (Wu et al. 2016; Vaswani et al. 2017). This technological evolution has not only disrupted translation services provision (Cid et al. 2020), but has also shaped professional translation workflows, impacting expectations, cost structures, and quality assurance protocols (ELIS Research 2025).

The recent emergence of large language models (LLMs) marks an additional disruptive shift in the MT landscape. These models—trained on massive multilingual corpora and capable of sophisticated semantic reasoning—have demonstrated remarkable performance on a range of natural language understanding and generation tasks, including translation, among others (Brown et al. 2020). Early studies suggest that, in some contexts and for some language pairs, LLMs may rival or surpass traditional NMT systems, particularly in terms of contextual adaptation, stylistic control, and zero-shot generalization (Gao et al. 2023; Hendy et al. 2023; Briva-Iglesias et al. 2024).

In parallel, LLM-based AI agents—autonomous or semi-autonomous systems capable of decomposing tasks, invoking external tools, and reasoning through iterative workflows—have begun to demonstrate considerable success in fields such as software engineering (Qian et al. 2024), customer support (Li, Zhang, and Sun 2023), and data analysis (Wang et al. 2023). While promising, the application of such AI agents to MT remains nascent and underexplored. These AI agent workflows introduce a novel paradigm: one that operationalizes translation as a multi-step, modular, and potentially collaborative process.

This paper aims to provide an empirical comparison of five state-of-the-art MT paradigms: a standard NMT system (Google Translate), a general-purpose LLM (GPT-4o), a reasoning-enhanced LLM (o1-preview), and two AI agent systems—one sequential (s-agent), the other iterative (i-agent). These systems are evaluated across three key dimensions: (i) quantitative quality (via automatic metrics), (ii) qualitative quality (via human expert evaluation), and (iii) cost-efficiency (measured in token-based computational expenditure). The evaluation is conducted in three language pairs: English-Spanish (between high-resource languages), English-Catalan (high to medium-resource languages), and English-Turkish (high to medium-resource languages). By isolating architecture-specific effects while holding the underlying model constant where applicable via the same set of prompts, we provide a robust assessment of the relative strengths and limitations of these paradigms. To guide our investigation, we formulate the following research questions (RQ):

- RQ1: How do LLM-based AI agent workflows compare to standard NMT and LLM translation systems in terms of translation quality, both quantitative and qualitative?
- RQ2: What are the cost-performance trade-offs associated with each translation paradigm, and do higher computational costs correlate with higher quality?

By addressing these questions, this study contributes to the emerging body of literature on AI agents in MT and sheds light on the viability, scalability, and practicality of agent-based MT in real-world applications.

## 2. Literature Review

MT has undergone a series of paradigm shifts, evolving from rule-based and statistical methods to NMT. The advent of NMT, particularly models leveraging encoder-decoder architectures with attention mechanisms (Bahdanau et al. 2016; Vaswani et al. 2017), marked a significant leap forward in translation quality. NMT systems demonstrated superior fluency and semantic coherence over phrase-based statistical MT, and quickly became the industry standard across most major language service providers and platforms (Castilho et al. 2017). Despite their impressive performance, traditional NMT systems have limitations. These include poor performance in low-resource scenarios, lack of explicit world knowledge, and an often rigid handling of contextual and pragmatic nuance—particularly in domain-specific or culturally loaded content (Castilho et al. 2017; Kenny 2022). Moreover, their static nature and reliance on task-specific training pipelines constrain adaptability to new tasks or domains without further fine-tuning, which is resource intensive and costly, both economically and environmentally (Zhong et al. 2023).

The emergence of LLMs has redefined the limits of natural language processing and generation (Brown et al. 2020). These models, trained on vast multilingual corpora and capable of few-shot or zero-shot generalization, have demonstrated strong performance in translation tasks—sometimes even surpassing conventional NMT systems, especially in high-resource language pairs (Hendy et al. 2023). Unlike NMT systems, LLMs operate with broader context windows, can integrate meta-linguistic cues, and adapt to domain-specific instructions via prompt engineering (Gao et al. 2023; Sahoo et al. 2024). This flexibility enables higher customizability and contextual sensitivity, with the additional benefit of not requiring retraining for new tasks. Recent benchmarking efforts suggest that LLMs approach or match state-of-the-art NMT performance in many scenarios (Hendy et al. 2023; Briva-Iglesias et al. 2024).

While LLMs have made translation more powerful, they often function as monolithic systems. The notion of decomposing translation into modular steps using AI agents introduces a new design paradigm: AI agent-based translation workflows (Briva-Iglesias 2025). In such workflows, multiple specialized agents—each responsible for tasks such as translation, adequacy review, fluency refinement, and final editing—collaborate to iteratively produce high-quality outputs. AI agents, particularly those built upon LLMs, have demonstrated success in various domains requiring strategic reasoning and tool use, including software development (Qian et al. 2024), customer service (Li, Zhang, and Sun 2023), and data analysis (Wang et al. 2023). However, their application to MT remains underexplored. Cheng et al. (2024) provide a broader taxonomy of LLM-based intelligent agents, distinguishing between single-agent and multi-agent systems. They highlight the enhanced reasoning, memory, and planning capabilities of agents organized into collaborative frameworks. Multi-agent systems, in particular, allow for iterative refinement, feedback loops, and the integration of external tools—capabilities that could significantly benefit translation tasks involving high accuracy or compliance requirements.

Preliminary results from multi-agent systems (e.g., TransAgents by Wu et al. 2024; Ng [2024] 2025; Sin et al. 2025; Briva-Iglesias 2025) suggest that multi-agent workflows can outperform traditional MT and even vanilla LLM prompts in terms of perceived fluency, contextual alignment, and cultural nuance. Nonetheless, the evaluation of these studies often offered a limited human evaluation or cost-efficiency analyses, highlighting a need for more systematic and comparative research. Briva-Iglesias (2025) also outlines different architectures that simulate professional translation workflows using multi-agent systems, achieving promising results in the legal domain by mimicking the distributed roles seen in human translation teams.

However, it is worth stressing that the integration of LLMs and AI agents into MT cannot be divorced from broader reflections on automation, labour, and ethics. As the translation industry increasingly prioritizes speed and cost-efficiency (ELIS Research 2025), translation processes are becoming more digitised and "mathematicised" (Moorkens 2023), often at the expense of human agency. Algorithms now influence work availability, pricing structures, and quality control, steering the industry toward maximally profitable, yet potentially alienating, workflows (do Carmo and Moorkens 2022; do Carmo 2020; Carreira 2024).

This drive toward automation raises critical questions. Herbert et al. (2023) show that poorly designed automation can erode professional identity, diminish job satisfaction, and deskill expert practitioners. Classic human factors research warns against "automation complacency"—overreliance on automated systems without adequate oversight (Parasuraman et al. 2000). Floridi et al. (2021) emphasize that automation in critical contexts, such as law or healthcare, must be approached with extreme caution, underscoring the principle that certain responsibilities should not be delegated to machines.

In the MT context, scholars have recently argued for greater scrutiny of the design and deployment choices behind translation technologies, with the goals of developing, adopting and deploying a more human-centered MT approach that aims to augment language professionals (Läubli et al. 2022; O'Brien 2023; Briva-Iglesias 2024) by following the human-centered AI paradigm (Shneiderman 2022). Quality assessment, too, must evolve: as Doherty and O'Brien (2014) and Way (2018) point out, the purpose, perishability, and communicative intent of a text dictate the necessary degree of human involvement. Canfora and Ottmann (2018) go further in proposing risk-based frameworks to evaluate when and how automation should be deployed, and Pym (2025) proposes to consider risk management as a key measure for working and interacting with MT. The rise of AI agents in MT invites a reassessment of how much autonomy we grant to machine-led workflows, and whether such delegation serves the interests of translation quality, professional well-being, and societal accountability. As Dignum (2020, 216) notes, the goal of AI should not be to replace humans, but to ensure "safe, beneficial, and fair" outcomes.

Given the technical, economic, and ethical stakes, the evaluation of emerging MT systems must move beyond automatic scores alone. Metrics such as BLEU, COMET, and chrF2 offer useful, if narrow, proxies for translation quality (Kocmi et al. 2021). They fail, however, to capture dimensions like pragmatic adequacy, stylistic fit, or regulatory compliance (Ibid.). Human evaluation remains the go-to methodology, particularly when conducted by domain experts and contextualized within real-world use cases (Läubli et al. 2020). Equally important is the cost-efficiency dimension, that has been substantially neglected to date. As token-based billing models become standard and AI prices start to be commoditised, especially in LLM and agent-based systems, understanding the trade-off between quality and economic feasibility is paramount. This paper contributes to these discussions by offering a multidimensional evaluation of state-of-the-art MT systems—including two novel AI agent workflows—across both quantitative and qualitative axes, and with explicit attention to costs and broader impacts.

3. Methodology

This section details the step-by-step methodology used in the paper.

3.1. MT systems and translation generation/prompt engineering

This study evaluates five distinct MT systems representing diverse translation paradigms, ranging from traditional NMT to LLM-based MT and novel multi-agent-based workflows. MT outputs were generated using different prompt strategies and methods depending on each system. The evaluated MT systems are as follows:

- Google Translate (GT): A widely adopted, state-of-the-art NMT system chosen due to its robust multilingual support and benchmark performance. According to Hickey (2023), it is the most widely used NMT system worldwide. Outputs were obtained via direct web interface to Google's translation service.
- GPT-4o (model gpt-4o-2024-08-06): A state-of-the-art LLM utilizing standard decoder-only approaches without additional reasoning or iterative mechanisms.  Translations were generated via ChatGPT's web interface.
- o1-preview: A reasoning-enhanced LLM specifically designed to incorporate advanced semantic coherence and contextual alignment in translation tasks. Translations were generated via ChatGPT's web interface.
- Sequential multi-agent system (s-agent): A structured workflow comprising three specialized agents—translator, reviewer, and editor—operating sequentially. This means that the first step had to be completed for the second step to start. The s-agent was implemented using Andrew Ng's framework and code[1] (Ng [2024] 2025).
- Iterative multi-agent system (i-agent): An adaptive multi-agent framework employing the same agent roles as the sequential system but with an iterative workflow, allowing up to three revision cycles aimed at optimizing translation quality. This system was built by using the CrewAI library[2] and an evaluator-optimizer AI agent workflow (Briva-Iglesias, 2025). Both multi-agent frameworks were powered by the GPT-4o-2024-08-06 model, using exactly the same prompts and underlying MT model. Consequently, any differences in resulting quality are attributed exclusively to their distinct architectures, rather than differences in MT models or prompts.

3.2. Text selection

Two distinct texts were selected to test translation systems under realistic and replicable conditions. On the first hand, a legal contract: extracted from a previously published study (ANONYMISED) containing 537 words and a type-token ratio of 0.305. The complexity of this text, characterized by specific terminology, numerical data, and long syntactic structures, ensures rigorous assessment of domain-specific translation performance in the legal domain. On the second hand, a news document, extracted from Andrew Ng's publicly available GitHub repository, comprising 116 words and a type-token ration of 0.654. This text was chosen for its contemporary relevance and to ensure methodological replicability and validation against an openly accessible resource.

---

[1] https://github.com/andrewyng/translation-agent (last access: 30.04.2025)
[2] https://github.com/crewAIInc/crewAI (last access: 30.04.2025)

## 3.3. Human Evaluation

Professional human evaluators with over 10 years of translation expertise were recruited to conduct comprehensive quality assessments of translation outputs following established industry standards (Läubli et al. 2020). The evaluation protocol consisted of two primary dimensions—fluency and adequacy—rated on a 4-point Likert scale:
- Adequacy: Evaluators assessed semantic fidelity, terminological precision, factual correctness, and contextual suitability relative to the source text. Ratings ranged from 1 (lowest adequacy) to 4 (highest adequacy).
- Fluency: Evaluators reviewed linguistic naturalness, readability, and coherence of the translated text. Ratings ranged from 1 (lowest fluency) to 4 (highest fluency).

## 3.4. Cost Assessment

Given the growing economic and environmental importance of computational resources in translation, we included a token-based cost analysis. Due to the difficulty resulting from the ongoing competition between large providers of LLMs, the dollar costs of each translation tasks quickly become obsolete. In order to solve this problem, we calculated the input and output tokens generated by each model and provided a link to the up-to-date pricing pages of the providers (late April 2025). The metric of the total token count allows for clear benchmarking of economic efficiency alongside translation quality, enabling analysis of cost-performance trade-offs in professional translation contexts.

The methodological rigour and transparency of this study not only facilitate reproducibility but also contribute significantly to ongoing discourse on the application, performance, and sustainability of emerging MT technologies.

## 4. Results

This section presents the results of the paper, which are first analysed by assessing translation quality with automatic evaluation metrics (AEMs), and then via human evaluation (HE).

## 4.1. Automatic Evaluation Metrics

Table 1 shows the AEMs' results. The best-performing system for each AEM and language pair is highlighted in **bold**, while the second-best performing system is underlined. This visual distinction allows for an immediate understanding of each system's comparative strengths.

**Table 1.** Automatic evaluation results for EN-ES, EN-CA and EN-TR pairs. Bold figures highlight the best performing system; underlined figures indicate the second best performing system.

| T. Language | System | COMET↑ | BLEU↑ | chrF2↑ | TER↓ |
|---|---|---|---|---|---|
| Spanish | GT | **85.2** | **29.3** | **57.1** | **54.4** |
| Spanish | GPT-4o | 82.5 | 28 | 55.6 | 55.3 |
| Spanish | o1--preview | 83.6 | **29.3** | 56.1 | **54.4** |
| Spanish | s-agent | 83.2 | 27.4 | 55.3 | 56.1 |
| Spanish | i-agent | 82 | 27.3 | 56.9 | 56.9 |
| Catalan | GT | **84.6** | 23.6 | 52.9 | 60.6 |
| Catalan | GPT-4o | 81.9 | 23.5 | 50.3 | 61.4 |
| Catalan | o1-preview | 83.3 | 24.1 | **53** | **58.7** |
| Catalan | s-agent | 82.1 | **24.3** | 52.1 | 60.6 |
| Catalan | i-agent | 82.9 | 20.8 | 50.3 | 61.8 |
| Turkish | GT | 90.4 | **29.8** | **62.5** | 60.9 |
| Turkish | GPT-4o | 90.1 | 22.5 | 54.2 | 64.5 |
| Turkish | o1--preview | 90.4 | 29.2 | 58.6 | **59.6** |
| Turkish | s-agent | **91** | 24.7 | 57.3 | 61.6 |
| Turkish | i-agent | 86.7 | 18.9 | 49.2 | 74.6 |

Across a total of 12 comparisons (four metrics x three languages), GT achieves the highest number of **first-place results**, ranking first in 7 out of 12 cases, with some cases resulting in tied positions with other systems. The reasoning-enhanced model o1-preview secures 4 first-place rankings, consistently competing with GT and, in some cases, equaling its performance. The s-agent system attains 2 first-place results, and neither GPT-4o nor the i-agent system achieve any first-place results across the twelve comparisons.

When considering second-best performers, o1-preview emerges as the most consistent competitor, appearing as the second-best system in 6 out of 12 comparisons, while GT only attains this score 3 times. GPT-4o attains 2 second-place positions and generally ranks below GT and o1-preview. The i-agent system also secures 1 second-best ranking, while there is no single case where the s-agent system obtains this score.

These results point to several important interpretations. First, GT's dominance in the automatic metrics is consistent with expectations, given its longstanding optimization for surface-level metrics such as BLEU and COMET. Its architecture, heavily trained on vast multilingual datasets and fine-tuned for general-purpose translation, provides strong baseline performance across a variety of tasks and languages. GT remains a system that maximizes n-gram overlap and semantic adequacy in line with traditional automatic evaluation paradigms (Kocmi et al. 2021).

Second, the competitive performance of o1-preview demonstrates the emerging power of reasoning-enhanced LLMs (Liu et al. 2025). o1-preview often narrows or closes the gap with traditional NMT systems, according to AEMs; however, these results need to be cross-checked with HE.

Third, the relatively lower performance of both multi-agent systems across AEMs underscores a critical insight: AI agent workflows, despite their theoretical advantages in enhancing quality through multiple review stages, are not necessarily optimized for AEM success. Indeed, token-based iterative revision may improve some dimensions of translation quality invisible to AEMs, such as terminological coherence or stylistic appropriateness, but not necessarily n-gram precision or surface semantic matching. This needs further empirical validation.

Overall, these findings suggest that, while AI agent workflows and advanced LLMs offer promising innovations, they do not automatically outperform traditional NMT systems like GT under standard AEMs, which continue to favour architectures optimized for direct translation outputs that maximize surface correspondence with reference translations (Kocmi et al. 2021). This reinforces the notion that a purely metric-driven assessment may obscure deeper qualitative differences, especially in translation contexts where nuances of style, pragmatics, and domain-specific accuracy are critical. The results thus highlight the necessity of complementing automatic evaluations with human-centered assessments to capture the full potential and limitations of emerging MT paradigms.

4.2. Human Evaluation

HE provides more profound insights into the comparative quality of the MT outputs, capturing linguistic and pragmatic dimensions that AEMs often fail to fully reflect (Freitag et al. 2021). The HE results for the three language pairs under study are provided in detail in this section. Table 2 provides a snapshot of the fluency and adequacy results and, again, the best-performing system per criterion is highlighted in **bold**, and the second-best performing system is <u>underlined</u>.

**Table 2.** Human evaluation results for EN-ES, EN-CA and EN-TR pairs on a scale of 4. Bold figures highlight the best performing system; underlined figures indicate the second best performing system.

| Target Language | System | Fluency | Adequacy |
| --- | --- | --- | --- |

|         | System | | |
|---------|--------|---|---|
| Spanish | GT | <u>3.67</u> | 3.81 |
|         | GPT-4o | 3.64 | 3.83 |
|         | o1-preview | 3.64 | **3.92** |
|         | s-agent | 3.56 | 3.78 |
|         | i-agent | **3.72** | <u>3.86</u> |
| Catalan | GT | 3.33 | 3.56 |
|         | GPT-4o | 3.39 | 3.67 |
|         | o1-preview | **3.69** | **3.94** |
|         | s-agent | 3.44 | <u>3.75</u> |
|         | i-agent | <u>3.61</u> | 3.72 |
| Turkish | GT | 3.56 | <u>3.64</u> |
|         | GPT-4o | <u>3.64</u> | 3.5 |
|         | o1-preview | **3.69** | **3.86** |
|         | s-agent | 3.58 | <u>3.64</u> |
|         | i-agent | 3.44 | 3.56 |

Across the six HE comparisons (two scores x three languages), the reasoning-enhanced model o1-preview emerges as the most frequent best-performing system, achieving 5 first-place scores. Iterative agent workflows (i-agent) secures 1 first-place result, while the remaining systems do not achieve any best-performing system score in our HE.

Regarding second-best performances, the distribution is more balanced. Both multi-agent workflows (i-agent and s-agent) and GT achieve 2 second-best scores each, while GPT4o follows last with 1 second-best scores. These results suggest that while o1-preview dominates in absolute quality according to human judgment, multi-agent workflows and NMT also perform competitively.

Turning to a more granular analysis, language-specific patterns reveal additional nuances. In Spanish, o1-preview emerges as the highest-scoring system in terms of adequacy, with a score of 3.92, while the iterative agent system (i-agent) secures the highest fluency score at 3.72, closely followed by o1-preview and GPT-4o. Interestingly, although GT and GPT-4o perform robustly, their scores remain slightly lower across both criteria compared to the reasoning-enhanced and multi-agent workflows. The i-agent system, despite achieving strong fluency scores, demonstrates marginally lower adequacy compared to o1-preview. These results suggest that reasoning-enhanced LLMs and iterative multi-agent workflows can match or slightly surpass traditional MT systems in capturing the semantic fidelity and stylistic smoothness expected by professional translators, at least in a high-resource language such as Spanish.

In Catalan, o1-preview also stands out, attaining the highest adequacy score (3.94) and the highest fluency score (3.69) among the MT systems evaluated. Iterative and sequential agent systems perform competitively in Catalan as well, with iterative agents achieving high fluency (3.61) and sequential agents scoring strongly in adequacy (3.75). By contrast, GT and GPT-4o record lower scores, reflecting greater challenges in adequately handling the morphosyntactic complexity of Catalan. These results reinforce the hypothesis that multi-agent workflows and advanced reasoning architectures provide greater adaptability and domain sensitivity, even in medium-resource languages.

In Turkish, a different pattern emerges. Here, o1-preview once again leads in both adequacy (3.86) and fluency (3.69), demonstrating the robustness of reasoning-enhanced approaches even in a typologically distant and morphologically rich language. The sequential agent system (s-agent) performs moderately well, maintaining relatively high adequacy and fluency scores, whereas the iterative agent system (i-agent) struggles significantly, recording the lowest scores in both dimensions. GPT-4o and GT deliver middling performances, neither matching the semantic coherence nor the stylistic naturalness exhibited by o1-preview. These results suggest that in languages like Turkish, where structural and syntactic differences from English are substantial, more sophisticated semantic reasoning is required to achieve high-quality translation outputs. In this line, a closer examination of the s-agent and i-agent results in Turkish show that both systems have multiple incorrect term translations, inconsistent terms across the document and grammatical issues. The i-agent tended to deviate from the source text and had more mistranslated sentences.

### 4.3. Token Generation and Costs

In addition to quality considerations, an increasingly crucial aspect of MT workflows is their economic and computational cost (Zhong et al. 2023). As token-based billing models become the industry norm for LLMs and AI agent systems, understanding the trade-off between translation quality and computational resource expenditure becomes essential for assessing the viability of emerging approaches. This section provides an analysis of the input and output (I/O) token counts required for each system to generate translations across the three language pairs studied. These figures serve as proxies for both economic cost and environmental impact.

**Table 3.** Total number of tokens used to generate each translation divided by input and output (I/O) token. Price is shown in two forms: first, by using a per million (M) token usage (standard metric in industry); second, by measuring the cost of translating our selected texts (our MT output).

| Target Language | System | Total Tokens (I/O) | USD Price/M (April 2025) | USD Price for our MT output |
|---|---|---|---|---|
| Spanish | GT | 1930 (857+1073) | $20 (chars.) | $0.038 |
| | GPT-4o | 2091 (865+1226) | $2.5 (I)/$10 (O) | $0.014 |
| | o1-preview | N/A | $15 (I)& $60 (O) | N/A |
| | s-agent | 10607 (7177+3430) | $2.5 (I)/$10 (O) | $0.052 |
| | i-agent | 29325 (13817+15508) | $2.5 (I)/$10 (O) | $0.189 |
| Catalan | GT | 2017 (857+1160) | $20 (chars.) | $0.040 |
| | GPT-4o | 2026 (866+1160) | $2.5 (I)/$10 (O) | $0.014 |
| | o1-preview | N/A | $15 (I)& $60 (O) | N/A |
| | s-agent | 11768 (7808+3960) | $2.5 (I)/$10 (O) | $0.059 |
| | i-agent | 39415 (21797+17618) | $2.5 (I)/$10 (O) | $0.231 |
| Turkish | GT | 2070 (857+1213) | $20 (chars.) | $0.041 |
| | GPT-4o | 2079 (865+1214) | $2.5 (I)/$10 (O) | $0.014 |
| | o1-preview | N/A | $15 (I)& $60 (O) | N/A |
| | s-agent | 11562 (7681+3881) | $2.5 (I)/$10 (O) | $0.058 |
| | i-agent | 29638 (13794+15844) | $2.5 (I)/$10 (O) | $0.192 |

The results demonstrate a clear difference in token consumption. GT, which operates on a character-based pricing model, maintains the lowest token-equivalent footprint, with approximately 1900 to 2100 tokens required for our assessed documents. GPT-4o, representing standard LLM prompting, slightly exceeds GT's token count, ranging between 2000 and 2100 tokens. These differences are minor and suggest that single-pass LLM translation, in its most basic form, remains relatively economical in computational terms.

In contrast, multi-agent workflows incur significantly higher costs. The sequential agent system (s-agent) consumes between 10,000 and 12,000 tokens to translate our sample documents, depending on the language pair. The iterative agent system (i-agent) exhibits an even steeper increase, requiring between 29,000 and 39,000 tokens to complete the translation of our sample texts. This dramatic escalation reflects the multiple back-and-forth exchanges, revisions, and internal evaluations inherent to multi-agent architectures, particularly those that involve iterative refinement cycles. The iterative agent, for instance, can generate token counts up to fifteen times greater than that of the original input, depending on the complexity of the source text and the number of revision rounds triggered.

Due to the lack of API access at the time of evaluation, it was not possible to obtain precise token counts for o1-preview. Nonetheless, given the presence of additional reasoning layers and expanded contextual memory windows, it is reasonable to infer that o1-preview's token consumption would be higher than that of GPT-4o, though still substantially lower than that of the multi-agent systems.

These findings have important implications. From an economic perspective, the high token consumption of AI agent workflows translates into significantly higher operational costs. Even though token prices may decrease over time due to market competition, the multiplicative increase in tokens suggests that agent-based MT solutions could become prohibitively expensive for large-scale deployments unless carefully managed. From an environmental standpoint, the computational overhead associated with generating tens of thousands of tokens per document raises concerns about the carbon footprint and sustainability of multi-agent MT approaches, echoing broader critiques of AI scalability and energy consumption.

The key question, therefore, is whether the observed improvements in translation quality—particularly in terms of adequacy, domain-specific alignment, and cultural nuance, if any—justify the steep increase in cost and resource usage. While multi-agent workflows show promise in specific scenarios, particularly for high-stakes domains requiring precision and iterative review, their broad adoption may be limited unless substantial efficiency gains can be achieved.

Overall, the cost analysis highlights a critical trade-off that must be addressed in the future development of AI agent workflows for translation. Innovations that reduce token overhead without compromising quality—such as more efficient coordination strategies, selective activation of agents, or hybrid models that combine single-pass translation with targeted agent intervention—are likely to be necessary to ensure that the benefits of these new paradigms are both economically and environmentally sustainable.

5. Discussion

The results of this study contribute to a rapidly evolving conversation about the future of MT, particularly in light of recent advances in LLMs and AI agent workflows. Compared to traditional NMT systems, such as GT, both reasoning-enhanced LLMs and multi-agent approaches demonstrate significant potential to reshape how translation is performed, according to our HE. Yet, this promise is accompanied by important limitations, costs, and design trade-offs that merit critical reflection.

The HE findings indicate that reasoning-enhanced LLMs, such as o1-preview, systematically outperform traditional NMT systems across multiple dimensions of translation quality. This aligns with recent scholarly work on reasoning LLMs for MT (Liu et al. 2025). However, while o1-preview consistently delivered the highest adequacy and fluency scores, the multi-agent workflows also yielded strong results, even though these results were more variable depending on the language pair and the evaluation criteria. The multi-agent systems performed comparably to o1-preview in Spanish and Catalan, but underperformed notably in Turkish, highlighting that multi-agent workflows' advantages are highly language-sensitive.

These findings echo and extend prior work on LLM-based MT systems. In particular, they support the hypothesis that reasoning layers can mitigate some of the limitations observed in standard NMT and single-pass LLM outputs, such as limited contextual adaptation and stylistic incongruence (Gao et al. 2023; Hendy et al. 2023). However, they also suggest that reasoning alone is not a panacea: while it improves semantic adequacy, it can introduce lexical and syntactic divergence from human references, which may be penalized under traditional AEMs such as COMET or BLEU. This phenomenon, already hinted at in Briva-Iglesias et al. (2024) and further corroborated by preliminary findings from Cheng et al. (2024), points to an emerging misalignment between existing evaluation frameworks and the qualitative strengths of reasoning-driven MT. Should we move beyond AEMs and put a higher emphasis on strengthening HE or developing newer evaluation metrics that do not jeopardise LLM-based MT outputs?

While our study focuses on quantitative human evaluation of the outputs from multi-agent workflows, a fine-grained deeper qualitative evaluation of them can provide further understanding about their overall translation behavior and expose their strengths and weaknesses, which we plan to concentrate on a future study.

Multi-agent workflows offer an additional and distinctive contribution to this landscape. By operationalizing translation as a multi-step, modular process involving task-specific specialization and iterative refinement, they introduce new forms of control and potential quality assurance mechanisms into MT. In scenarios where adequacy, terminological precision, or domain-specific fidelity are paramount—such as legal or healthcare translation—these workflows may better approximate professional human workflows than single-pass systems. Nonetheless, their technical complexity and substantial increase in computational and economic costs associated with agent-based architectures presents a major barrier to their broader adoption. Token consumption for multi-agent systems was found to be multiple orders of magnitude greater than that of baseline systems, raising significant concerns regarding scalability, affordability, and environmental impact.

Furthermore, the variability of AI agent workflow performance across language pairs suggests that these systems are not yet robust enough for universal deployment. Their success appears contingent on factors such as language typology, domain complexity, and the effectiveness of inter-agent communication protocols. We acknowledge that, the iterative agent (i-agent) system that we developed had some technical limitations, and that the results could have been substantially better with a more adequate development. This finding resonates with critiques raised by Way (2018) and Canfora and Ottmann (2018) regarding the need for risk-based assessments in MT deployment, and suggests that careful evaluation of domain suitability will be crucial before multi-agent MT can be responsibly operationalized at scale.

An important takeaway from this study, therefore, is that no single MT paradigm currently represents a definitive solution. Traditional NMT systems remain cost-effective and reliable for general-purpose translation; reasoning-enhanced LLMs offer clear improvements in semantic depth and stylistic control; and multi-agent workflows hold the promise of domain-sensitive, modular quality enhancement, albeit at a high computational cost. The selection of an MT approach must therefore be context-dependent, weighing not only output quality but also cost-efficiency, sustainability, and the specific demands of the use case.

Finally, the results also reinforce a broader methodological implication: evaluation methodologies in MT research must evolve to accommodate the emerging diversity of MT systems. Traditional AEMs, while still useful for benchmarking, increasingly fail to capture critical qualitative dimensions such as stylistic nuance, pragmatic appropriateness, or human-likeness in reasoning. Multi-dimensional HE—including not only typical adequacy and fluency scores, but also augmented by cost analysis and domain-specific risk assessment—should become a standard practice in assessing the next generation of MT technologies.

## 6. Conclusion

This study provides one of the first systematic evaluations of AI agent workflows and reasoning-enhanced LLMs within the field of MT, comparing them against traditional NMT systems across three language pairs, multiple quality dimensions, and cost considerations.

The results confirm that recent advances in LLM reasoning capabilities offer tangible improvements over conventional NMT, particularly in terms of semantic adequacy, contextual sensitivity, and stylistic coherence. At the same time, they reveal that agent-based workflows introduce novel possibilities for translation quality control, especially through modular specialization and iterative refinement, although these gains are accompanied by substantial increases in computational and economic cost.

Our findings indicate that reasoning-enhanced LLMs, such as o1-preview, consistently outperform traditional MT systems and foundational LLM prompts in both human and automatic evaluations. Agentic workflows, while demonstrating potential in specific contexts, display significant variability across language pairs, suggesting that their effectiveness is highly contingent on task-specific factors. Furthermore, the high token consumption associated with multi-agent approaches raises important questions about their scalability and environmental sustainability, which must be carefully considered before widespread adoption.

Critically, the study highlights the limitations of relying exclusively on AEMs when assessing emerging MT technologies. As MT systems become more complex and human-like in their reasoning, traditional metrics such as BLEU and COMET increasingly fail to capture higher-order qualities such as pragmatic fit, domain sensitivity, and overall translation acceptability. Future evaluation frameworks must therefore incorporate more sophisticated human-centered methods, including error typology analysis, domain-specific risk assessment, and cost measurement.

Ultimately, this paper argues that no single translation paradigm—whether traditional NMT, standalone LLMs, or multi-agent workflows—offers a universally optimal solution. Instead, the choice of technology should be informed by the specific requirements of the translation task, including the acceptable trade-offs between quality, cost, speed, and ethical considerations.

As AI-powered translation continues to evolve, future research should focus on developing more efficient and robust multi-agent workflows, optimizing reasoning strategies for diverse language pairs, and designing hybrid systems that combine the strengths of different paradigms. Only by pursuing these lines of inquiry can we ensure that the next generation of MT technologies not only enhances communication across languages but does so in a way that is safe, fair, and sustainable.